\documentclass{article}

\PassOptionsToPackage{numbers, compress}{natbib}

\usepackage[preprint]{neurips_2024}

\usepackage{natbib}

\usepackage[table,xcdraw]{xcolor}
\usepackage[utf8]{inputenc} 
\usepackage[T1]{fontenc}    
\usepackage{url}            
\usepackage{booktabs}       
\usepackage{amsfonts}       
\usepackage{nicefrac}       
\usepackage{microtype}      
\usepackage{xcolor}         
\usepackage[pdftex]{graphicx}
\usepackage{subcaption}
\usepackage{amsmath}

\usepackage{enumitem}
\usepackage{wrapfig}
\usepackage{hyperref}       

\title{
From Symbolic Tasks to Code Generation: Diversification Yields Better Task Performers
}

\author{Dylan Zhang \\
    University of Illinois \\Urbana-Champaign \\
  \texttt{shizhuo2@illinois.edu} \\\And
  Justin Wang \\
        University of Illinois \\Urbana-Champaign \\
  \texttt{jw93@illinois.edu} \\
  \\\And
  Francois Charton \\
        Meta AI Research \\
  \texttt{fcharton@meta.com} \\
  }
\begin{document}

\maketitle

\begin{abstract}

Instruction tuning -- tuning large language models on instruction-output pairs -- is a promising technique for making models better adapted to the real world. Yet, the key factors driving the model's capability to understand and follow instructions not seen during training remain under-explored. Our investigation begins with a series of synthetic experiments within the theoretical framework of a Turing-complete algorithm called Markov algorithm, which allows fine-grained control over the instruction-tuning data.   Generalization and robustness with respect to the training distribution emerge once a diverse enough set of tasks is provided, even though very few examples are provided for each task. We extend these initial results to a real-world application scenario of code generation and find that a more diverse instruction set, extending beyond code-related tasks, improves the performance of code generation. Our observations suggest that a more diverse semantic space for instruction-tuning sets greatly improves the model's ability to follow instructions and perform tasks. 

\end{abstract}

\section{Introduction}

The rapid advance of large language models (LLMs) is one of the most exciting recent developments in artificial intelligence. LLMs, pre-trained on large text corpora, have demonstrated impressive generalizable reasoning capabilities and can achieve remarkable performance across a broad set of tasks, ranging from natural language comprehension~\cite{wang2020superglue} and generation~\cite{brown2020language} to mathematical reasoning~\cite{cobbe2021gsm} and programming~\cite{chen2021codex}. These models have shown promise in performing real-world tasks in various applications and business solutions. One fundamental pillar of such success lies in the capabilities of these models to perform tasks through generalizable reasoning.

\emph{Instruction tuning} is a popular approach to unlock the capabilities of these models originally trained on next-token-prediction objectives to understand instructions and perform reasoning to complete tasks. By training models on pairs of instructions and expected outcomes, instruction tuning teaches LLMs to perform specific tasks, thus enabling them to address real-world problems and seamlessly interact with humans. In practice, however, fine-tuning data is limited, and instruction tuning can only focus on a limited set of tasks. Its success is therefore critically dependent on the model's ability to generalize beyond its fine-tuning instructions to \emph{unseen tasks} not encountered during training. Several factors influence this generalization: the size of the fine-tuning sample, the diversity of the instruction sets, and the quality of the annotations. Yet, there is little systematic research on their relative impact.

Our work aims to fill this gap by proposing \textbf{two contributions} to the study of generalization in instruction tuning. First, we propose a systematic analysis of the \textbf{impact of instruction diversity} by focusing on a simple yet important symbolic task: \textbf{string rewrites}. This basic setting enables us to exercise fine control over the factors that may affect generalization and to demonstrate the importance of instruction diversity. To highlight the broader applicability, we describe this task in terms of a Markov algorithm, a classic Turing-complete model, ensuring rigorous examination of string replacements. Second, we extend this analysis to \textbf{a real-world application: code generation}, and show that fine-tuning on an instruction set that extends beyond coding tasks significantly improves performance.

Our \textbf{main findings} are:
\begin{enumerate}[nosep]
\item Instruction diversity is the main driver of generalization. Models trained on a diverse set of instructions generalize better, even when the number of examples per instruction is small.
\item The semantic diversity of instructions matters, together with the number of instructions.
\item Instruction diversity improves model robustness and can compensate for the adverse impact of non-uniform fine-tuning distributions.
\end{enumerate}

We demonstrate that generalization and robustness with respect to the training distribution emerge once a diverse enough set of tasks is provided, even though very few examples are provided for each task. These initial results are extended to a real-world application scenario of code generation, where we find that a more diverse instruction set, extending beyond code-related tasks, improves performance. Our observations suggest that a more diverse semantic space for instruction-tuning sets greatly improves the model's ability to follow instructions and perform tasks. 
\section{Related works}

\textbf{Datasets for instruction-tuning.}
Many datasets for instruction-tuning have been proposed. The best quality is achieved for sets collated by human 
annotators~\cite{khashabi2020unifiedqa,ye2021crossfit,sanh2022multitask,wang2022supernaturalinstructions,longpre2023flan,DatabricksBlog2023DollyV2,köpf2023openassistant}, but their size is constrained by the cost of annotation. Alternative methods, which use large language models to generate instruction sets, have been 
proposed~\cite{wang2023selfinstruct,honovich2022unnatural,alpaca,peng2023instruction,vicuna2023,xu2023wizardlm,köksal2023longform,kim2023cot}. They provide larger instruction sets, at the cost of reduced annotation quality. 

\textbf{Data curation for instruction-tuning.}
It is widely recognized that the quality of instruction-tuning datasets has a massive impact on the performance of fine-tuned models. Previous works acknowledged the contributions of several key factors.
Most research on the subject insist on the importance of the size and quality of the instruction sets~\citep{chung2022scaling,iyer2022opt,wang2023far}. 
Liang et al.~\citep{liang2024formatconsistency} point out the importance of consistent formats.
Several recent works~\cite{zhou2023lima,cao2023mining} suggest that models fine-tuned on carefully selected examples can achieve high performance with small datasets.

Various strategies for data curation have been proposed, focusing on instruction diversity, and the quality of answers ~\citep{zhou2023lima,cao2023mining,xu2023rethinking,li2024selective,liu2024selectit}. 
Several authors discuss the benefit of mixing tasks from different 
categories~\cite{longpre2023flan,iyer2022opt,bukharin2024data}. Closest to our work, Dong et al.~\citep{dong2024composition} discuss the impact of mixing general and domain-specific instructions, in order to achieve the best results with the smallest dataset.

\section{String replacements}

Our first synthetic task is \emph{string replacement}. The model is trained to apply a replacement rule $R$ (a pair of string letters, like $aa \to bac$) to some string of letters $I$ (e.g. $caaba$), resulting in an output string $O$ (e.g. $cbacba$). Rule $R$ is applied once only, to the leftmost occurrence of the rule. If rule $R$ be applied, the models returns the initial string $I$. For instance, applying rule $R: iss \to art $ to $I=mississipi$ yields $O= martissipi$, and applying $R$ to $I=canada$ yields $O=canada$.

Despite its simplicity, string replacement play a central role in theoretical computer science and formal logic. It is the basic operation in Markov Algorithms~\citep{markov54}, a Turing-complete model of computation. A Markov algorithm processes sequences of letters on a fixed alphabet by means of an ordered set of rewrite rules $R: (x_i \rightarrow y_i)_{i\in \{1,\dots, n\}},$ with $x_i$ and $y_i$ words over an extended alphabet.

\begin{figure}[ht]
     \centering
     \begin{subfigure}[b]{.38\textwidth}
         \centering
         \includegraphics[width=.98\columnwidth]{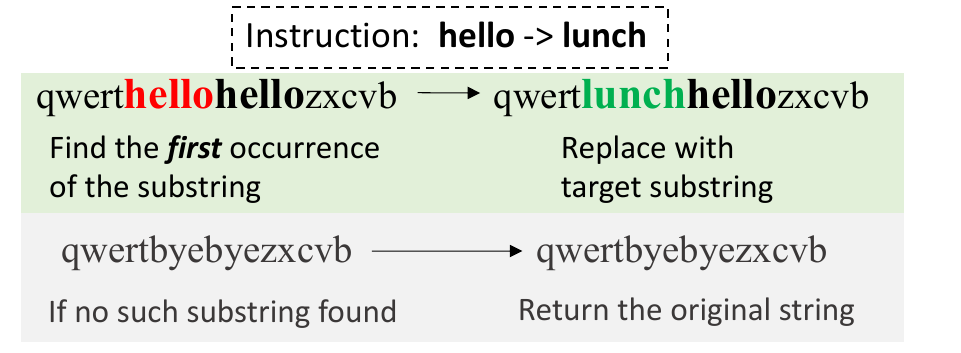}
         \caption{\small Our string-rewriting task set-up.}
         \label{fig:markov}
     \end{subfigure}
     \hfill
     \begin{subfigure}[b]{.6\textwidth}
         \centering
         \includegraphics[width=.98\columnwidth]{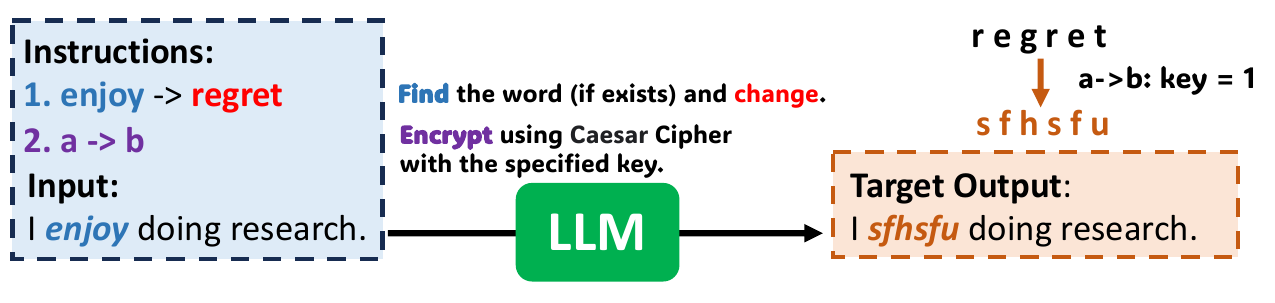}
         \caption{\small Encrypted Re-writing. Similar to the re-writing experiment, but the model needs to infer the rule of encryption based on the rule of shifting specified in the instructions and compute the encrypted word.}
         \label{fig:realworld}
     \end{subfigure}
     \caption{\small Illustration of our symbolic tasks in this paper.}
     \label{fig:symbolic_tasks}
\end{figure}
When processing a sequence $z_0$, the algorithms considers all rules in $R$ in order, finds the first one that applies to $z_0$ (say $x_i \to y_i$), and produces $z_1$ by replacing the leftmost occurence of $x_i$ in $z_0$ by $y_i$. The algorithms is then applied to $z_1$, iteratively producing $z_2, z_3, \dots z_n$, and terminates either when a special rule of the form $x_i \to \cdot$ is encountered, causing the model to return $z_n$, its last input string, or when no rules can be applied, in which case the algorithm is said to be blocked (and produces no output). Appendix~\ref{app:markov} provides examples of Markov algorithms.

Although very simples, Markov algorithms can be show to be Turing-complete: any finite computation can be implemented by a Markov Algorithm (this is Markov's thesis). As a result, a language model that can be trained to implement string rewrites could in theory become a universal computation machine (the practical difficulty, here, being the model capability to handle large number of rules, and its reliability over large numbers of rewrites). 

In this paper, we consider two tasks based on string rewrites:
\begin{enumerate}[nosep]
    \item apply rule $x\to y$ to a sequence containing $x$,
    \item apply rule $x\to y$ when the sequence contains $x$, return the input sequence when it does not. 
\end{enumerate}

Task $1$ is the base rewrite operation. Task $2$, referred to as ``no-op'' involves an additional decision step, on whether the rule is applicable. It is central to Markov algorithms, where the model must decide, at every step, which rule to apply.

In our experiments, model inputs and outputs are sequences of the lowercase letters $a\dots z$. Model inputs are triplets of strings, $(x,y,z)$ (separated by a special token), representing the rule $x \to y$ and the input sequence $z$. Model outputs are the strings $z'$ obtained by replacing the leftmost instance of $x$ by $y$, when $x$ is in $z$, or $z$ if $x$ is not in $z$. Our models uses a very limited vocabulary of $29$ tokens ($26$ lowercase letters, a special separator token, and beginning and end of sequence tokens).

\section{Experiments with string replacements}

In this first set of experiments, we train GPT-2~\cite{Radford2019gpt2} models with 256 dimensions, $6$ layers, and $4$ attention heads. The model is trained from scratch, and supervised, on a generated dataset of instruction/outcome pairs. The training details can be found in Appendix~\ref{task}.

The training sets include $S\times I$ input sequences, featuring $I$ different replacements rules (instructions), applied to $S$ different input sequences. Trained models are then tested on $10^5$ examples, all featuring unseen instructions. The goal of these experiments is to assess the relative impact of $S$ and $I$ for generalization to unseen instructions (measured by the accuracy of the trained model on the test set).

\begin{figure}[ht]
     \centering
     \begin{subfigure}[t]{.48\textwidth}
         \centering\includegraphics[width=\columnwidth]{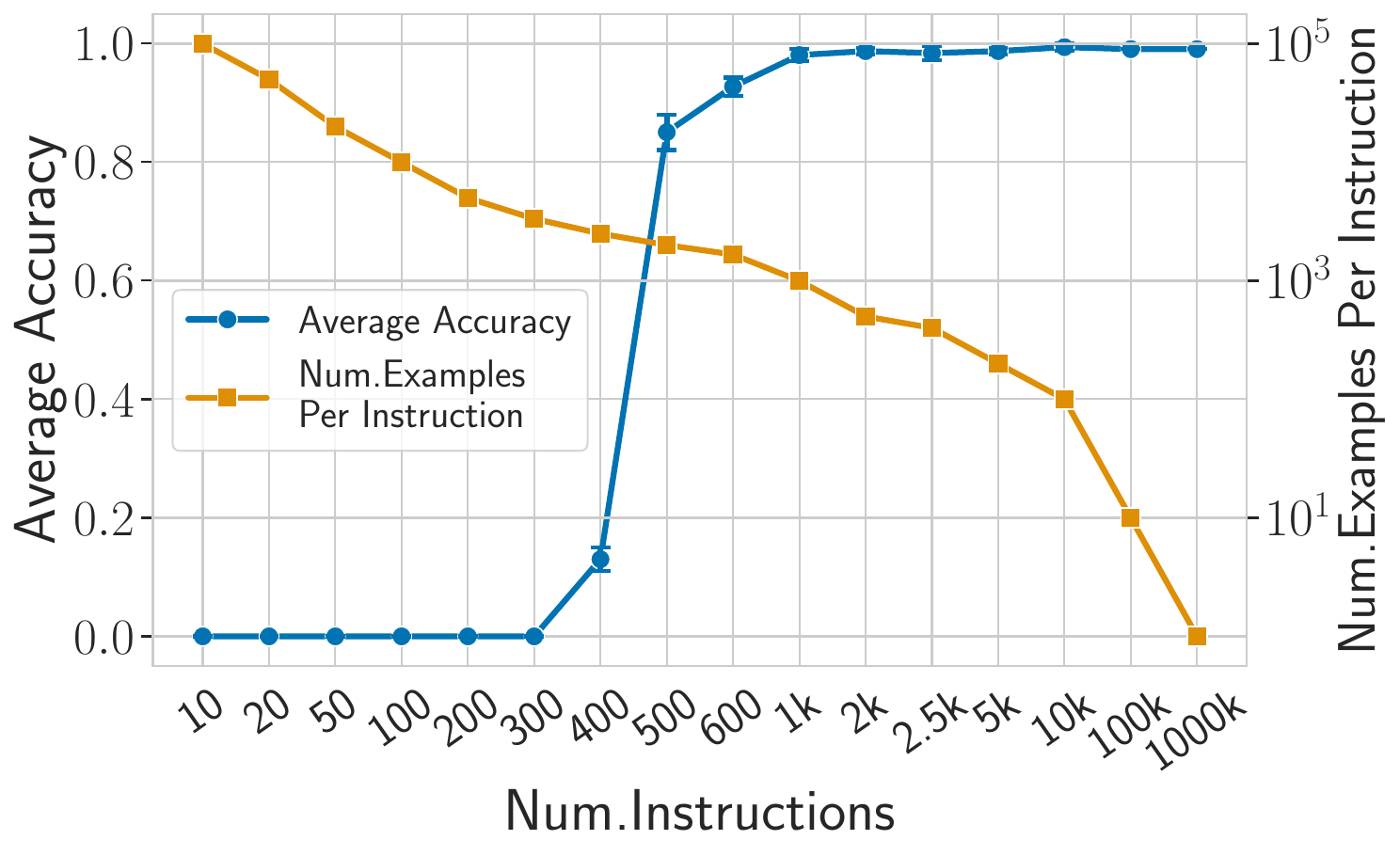}
         \caption{\small Re-writing accuracy against the number of instructions with a fixed-size training set.}
         \label{fig:one_occurence}
     \end{subfigure}
     \hfill
     \centering
     \begin{subfigure}[t]{.42\textwidth}
         \centering
         \includegraphics[width=\columnwidth]{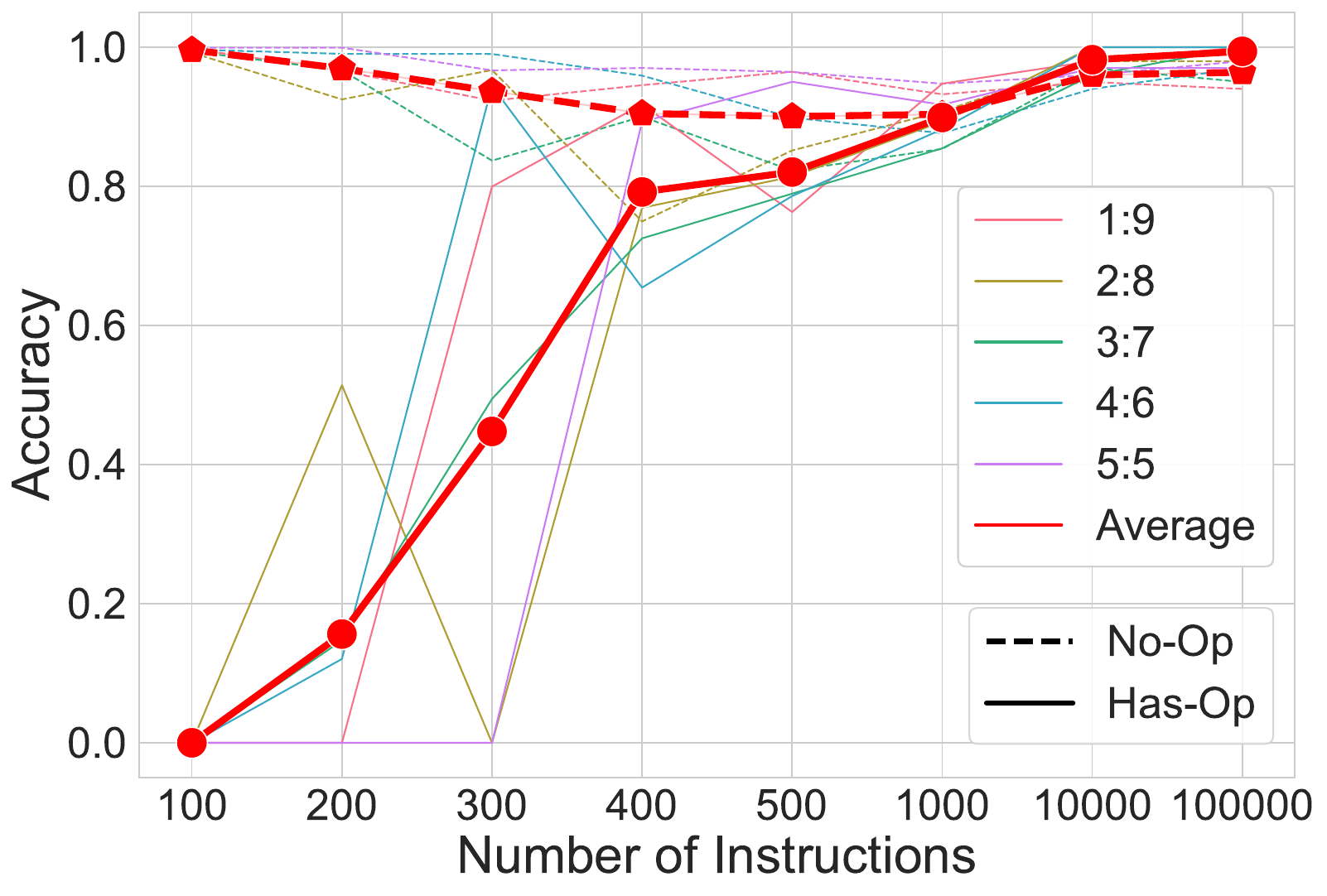}
         \vfill
         \caption{\small Rewriting with no-op situation included. }
         
         \label{fig:zero_one}
     \end{subfigure}
     \caption{\small Generalization versus the number of instructions during training. }
\end{figure}

\subsection{Instruction Diversity Is Decisive To Generalization}
Figure~\ref{fig:one_occurence} presents the generalization accuracy for models trained on a fixed budget of $I\times S = 10^6$ examples, as a function of the number $I$ of 
different instructions in the training set. The number of examples per instruction ($S$) decreases as $I$ increases. In these experiments, all input sequences feature at least one instance of the replacement rule: we are learning task 1.

The accuracy curve has a step shape. Models trained on less than $300$ rules (instructions) never generalize to unseen instructions, no matter the number of examples provided for each rule. On the other hand, models trained on $1000$ rules or more always generalize, even when each rule is featured in a handful of examples. A sharp phase transition happens around $400$ instructions.
We conclude that the number of different rules in the training set ($I$) is the key factor that allows the model to generalize to unseen instructions.



\subsection{Diversification Allows Generalization In Case-Based Reasoning Set-Up}
\label{sec:no_op}
In previous experiments, the model was trained to perform a specific task: replacing a substring that was always present in the input sequence. We now consider a more general and challenging setup involving a new task where some rules may not apply. In such cases, the model returns the input sequence unchanged. This new setup presents a two-step task: first, the model must decide whether the rule applies; second, it must either perform the replacement or copy the input.

This scenario represents a broader context for reasoning with LLMs. In LLM-agent frameworks, No-Ops are akin to decision points in complex environments where the agent must determine the relevance of actions based on the current context~\cite{huang2024understandingplanning,wang2024llmagent}..

To explore this, we introduce a third parameter in the training set: the frequency of “No-Ops” (instructions that cannot be satisfied), which we vary between 10\% and 50\%. The size of the training and test sets remains the same as in previous experiments.

Figure~\ref{fig:zero_one} presents the generalization accuracies of trained models, as a function of the number of instructions and the frequency of No-Ops. Accuracies for Ops and No-Ops unseen instructions are measured separately (these are, in fact, different tasks). Given the simplicity of No-Ops cases and their predominance in the data\footnote{Consider a dataset containing 100,000 data points, 10\% No-Ops, and 100 rules. No-Ops takes up 10,000 in total, $\sim$11$\times$ of any has-Ops.}, the model defaults to apply it when the number of instructions are small. Has-Ops cases, on the other hand, display the same patterns as in the previous experiments:  the model exhibits full generalization to unseen instructions once the training set features more than a given number of different instructions (around 400, here), below this number, the model performs the lazy operation of applying No-Ops to all inputs. Interestingly, the proportion of No-Ops in the training set seems to have little impact on generalization. 

Overall, our conclusions remain consistent with previous experiments, albeit with a slightly lower number of instructions needed for generalization (400 vs. 500). This demonstrates the effectiveness of diversification in more complex scenarios involving case-based reasoning.

\subsection{Imbalanced Distribution Is Still Effective In Driving Generalization} 
In previous experiments, instructions were evenly distributed between examples in the training set: in a training set of $1$ million examples, with $500$ different instructions, every instruction would be featured $2000$ times. Such a situation is unlikely to happen in real-world settings. In real-world training sets, some tasks will be much more common than others (due to the availability of fine-tuning data and the nature of the tasks). 

\begin{figure}[ht]
     \centering
     \begin{subfigure}[h]{.5\textwidth}
         \centering
         \includegraphics[width=.8\columnwidth]{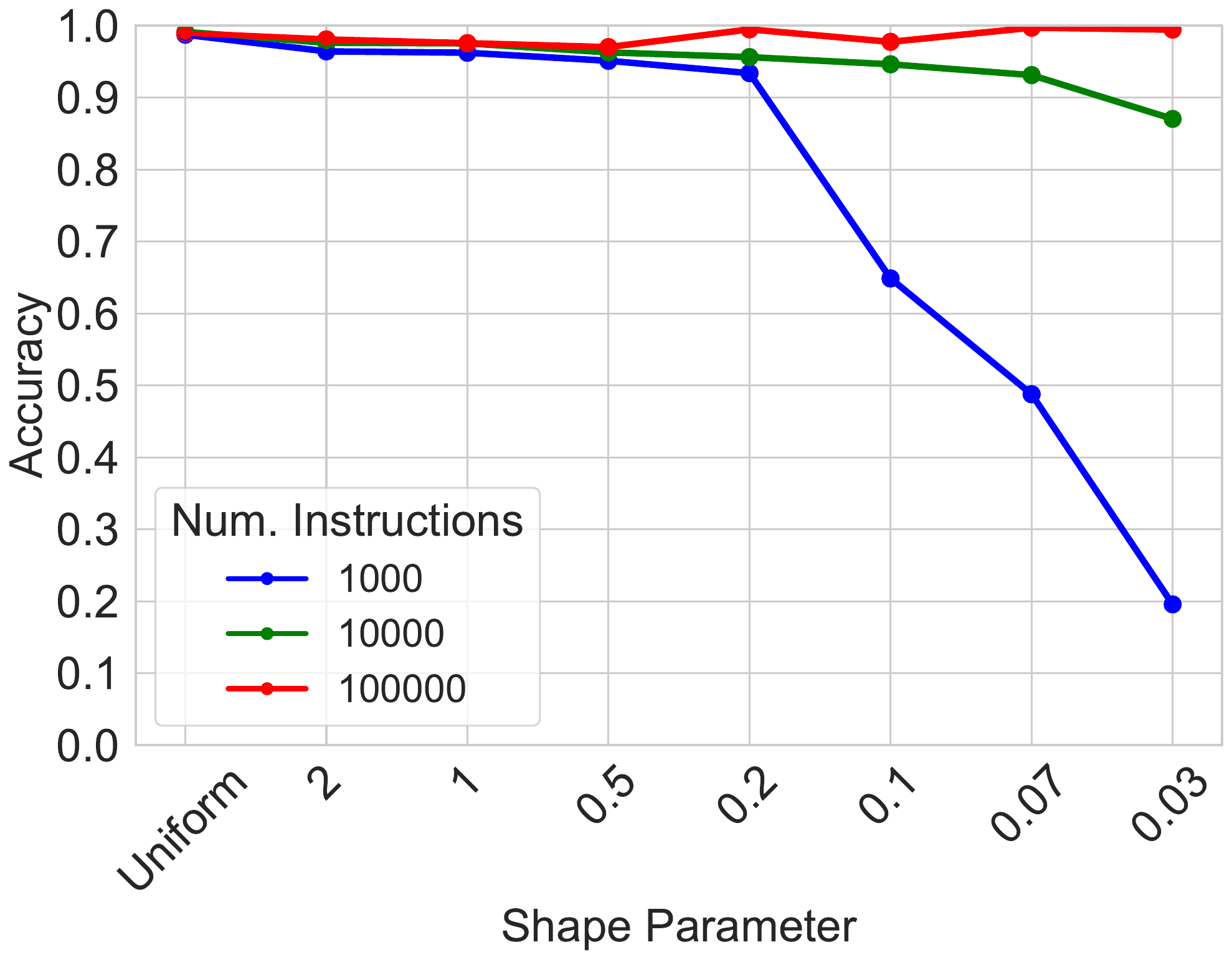}
         \caption{\small Effect of long-tail task distributions on model's generalization ability.}
         \label{fig:long_tail_distribution}
     \end{subfigure}
     \hfill
     \centering
     \begin{subfigure}[h]{.44\textwidth}
     \centering
         \includegraphics[width=.9\columnwidth]{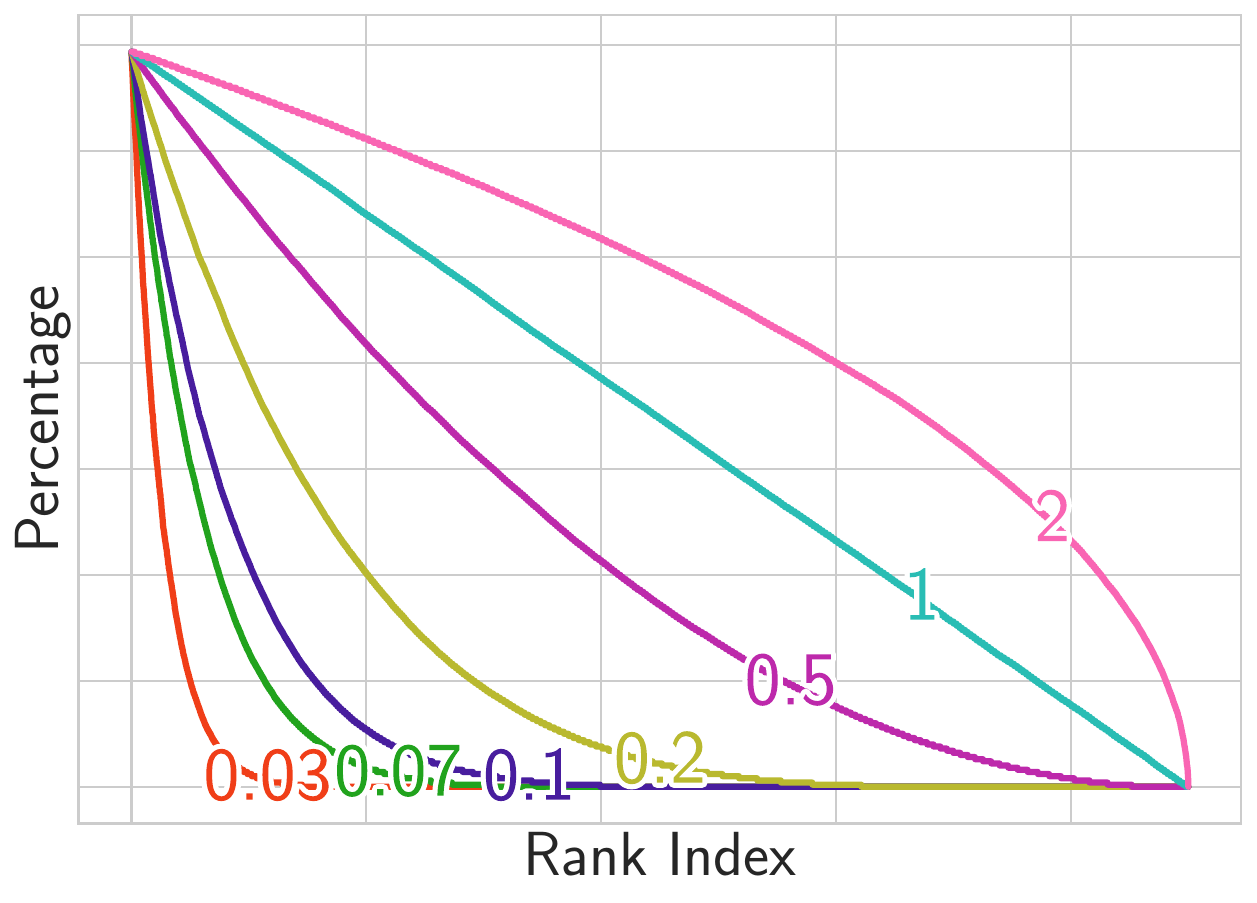}
         \caption{\small The sorted percentage of each instruction following power-law distribution with different shape parameters. The y-axis is the percentage of the rules in the training mixture. The x-axis is the ranked index (by proportion of examples) of instructions.} 
         \label{fig:distribution}
     \end{subfigure}
     \caption{\small Generalization versus number of instructions during training. }
\end{figure}

To investigate the impact of the distribution of instructions on generalization to unseen tasks, we generate datasets of 1,000, 10,000 and 100,000 different instructions,  
and distribute the number of examples per instruction according to a power law distribution with PMF $f(x) = \alpha x ^{\alpha-1}$ where $\alpha$ is the shape parameter. By varying the shape parameter of the power law, we can generate a distribution of examples that range from close to uniform, to extremely peaked as shown in Fig.~\ref{fig:distribution}.


Figure~\ref{fig:long_tail_distribution} presents model generalization as a function of the shape parameter of the power law, for training sets of $1$ million examples with 1,000, 10,000, and 100,000 instructions. Models trained on 10,000 different instructions or more prove to be robust with respect to the distribution of examples per instruction. For models trained on $1000$ instructions, generalization accuracy drop steeply when the shape parameter is larger than $0.2$. This observation nicely matches the result of previous experiments. In a training set of $1$ million examples, featuring $1000$ different instructions distributed according to a power law, instructions with a probability lower than $0.1\%$ are hardly visible. When the shape parameter falls below $0.2$, more than half of the instructions are below that threshold, and the model is effectively trained on less than $500$ instructions, the lower limit for generalization to unseen instructions.

\subsection{Semantic Diversification Boosts Task Performance}
\label{sec:diversity_semantics}
So far, instruction diversity has only been measured in terms of the number of instructions in the fine-tuning set. We now investigate the impact of semantic diversity in the instruction set. In the string rewrite setting, we constrain the semantic diversity of instructions by restraining the patterns that can appear in the rule substrings. In a semantically diverse set of rules, instructions would be randomly sampled from every possible sequence of lowercase letters. In a constrained setting, rule substrings must obey certain rules, like being composed of repeated letters, or repeated patterns, or having a palindromic structure.

Specifically, we experiment with three sets of semantic constraints: 
\begin{itemize}[nosep]
\item \textbf{repeated}: characters repeated $k$ times: $aaabbbccc$ for $k=3$,
\item \textbf{periodic}: patterns repeated $k$ times: $abcabc$ for $k=2$,
\item \textbf{mirror}: mirroring patterns repeated $k$ times: $abccbaabc$ for $k=3$,
\end{itemize}

and train models on sets of rules where both substrings obey the constraint (e.g. $aaabbb \to bbbccc$). Note that all sets of constrains depend on a parameter $k$, and that increasing $k$ makes the instruction set less diverse. To measure the impact of semantic diversity, we train models on instruction sets with large $k$, and test them on examples with low $k$.

First, we observe that models trained on one set of semantic constraints (e.g. all instructions repeated, or periodic, or mirror) with high values of $k$, \textbf{never generalize} to low values of $k$. Models ``overfit'' the high-$k$ semantic pattern. Training on a mixture of repeated and periodic instruction (with high $k$) bring no improvement: the model does not generalize to low $k$, for either constraint.

\begin{figure}[ht]
     \centering
         \includegraphics[width=.6\columnwidth]{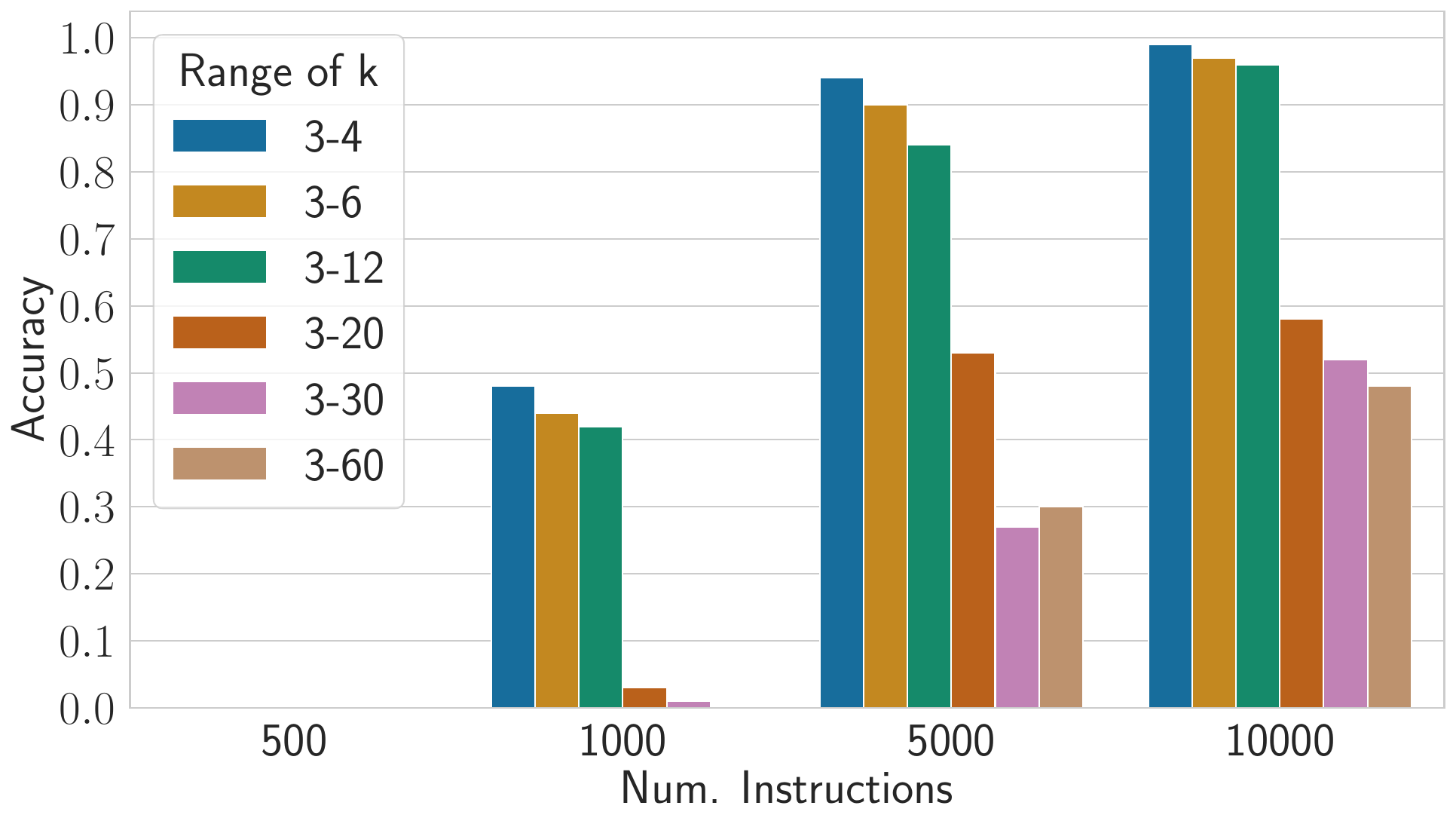}
         \caption{\small Model's performance on $k<3$ when trained on the three classes of restricted semantics as in~\ref{sec:diversity_semantics}. Models trained on 500 or less instructions never generalize to smaller k.}
         \label{fig:semantics}
\end{figure}
The situation changes when models are trained on a mixture of instructions from the three constrained sets~(Figure ~\ref{fig:semantics}). 
Models trained on large $k$ ($k$ between $3$ and $6$ for all three constraints) do generalize to small $k$ ($k<3$), and unconstrained instructions. As before the number of training instructions improves generalization. Models trained on only $500$ instructions do not generalize at all. Models trained on $5000$ achieve high accuracies. Finally, for a given number of instructions, training on more constrained sets, i.e. with larger $k$, makes generalization harder.

These experiments prove that instruction diversity goes further than having many different instructions in the fine-tuning dataset. Models trained on a diverse set of semantically constrained rules --the string rewrite equivalent of a diverse set of tasks -- generalize to broad sets of less constrained rules.

\subsection{Generalization Across Input Semantics}
\label{app:input_semantics}

In the previous section, we studied the impact on generalization of the semantic diversity of instructions. In this section, we focus on the diversity of the input sequences provided with the rules. Specifically, for a rule $x_i \to y_i$, applied to a sequence $s_i$, we study the impact of the distribution of $s_i$ in the training set, measured by the number  $\mathcal{O}$ of occurrences of $x_i$ in $s_i$ (always greater than $1$ for task $1$, this number of always larger than $1$). 

Table~\ref{tab:input_occurrence} presents the acccuracy of models trained on instruction sets with specific values of $\mathcal{O}$, on test sets with a uniform distribution o $\mathcal{O}$ between $1$ and $20$. For models trained on one value of $\mathcal O$, there is a  marked decline in generalization accuracy. And models with less than $500$ different instructions no longer learn to satisfactory levels. A more diverse instruction set, featuring with a wider range of $\mathcal O$ ($1,5,10,15$ and $20$) achieves much better generalization accuracy.

\begin{table}[ht]
\centering
\small 
\begin{tabular}{@{}l|llll@{}}
\toprule
& \multicolumn{4}{c}{Number of instructions} \\
Training set occurences $\mathcal{O}$  & 2000 & 1000 & 500  & 200  \\ 
\midrule
{1}                                                & 0.71 & 0.41 & 0.00 & 0.00 \\
{10}                                               & 0.88 & 0.71 & 0.00 & 0.00 \\
{15}                                               & 0.84 & 0.59 & 0.00 & 0.00 \\
{20}                                               & 0.73 & 0.28 & 0.07 & 0.00 \\
{1,5,10,15,20}                                     & \textbf{0.94} & \textbf{0.94} & \textbf{0.62} & \textbf{0.00} \\ \bottomrule
\end{tabular}
\small
\caption{\small Generalization across input semantics. $\mathcal{O}$ is the number of occurrences of the rule substring in the training set. Accuracies are measures on a test set with $\mathcal{O}$ uniformly distributed between 1 and 20}
\label{tab:input_occurrence}
\end{table}

This concludes our experiments on string replacement. The key takeaway is the undeniable importance of instruction diversity. To generalize to unseen rules, models must be trained on large ($\geq 500$) number of different instructions, featuring diverse rules and instructions.

\subsection{String replacements with pre-trained models}

Extending our results about string replacement to pre-trained models is not straightforward. After pre-training most LLM can perform string replacements with high accuracy, and it will be hard to tell whether their performance after fine-tuning is due to instruction tuning, or to their initial pre-training. To circumvent this, we introduce a more difficult task: encrypted rewrites. As in the string replacement task, the model is presented with a replacement rule $x \to y$ and a string to be replaced $s$, but the input also contains an encryption key $k$, and the model must replace $x$ by $E(y,k)$, the encrypted value of $y$ with key $k$: instead of replacing the string $s=s_1 x s_2$ by $s_1 y s_2$, the model replaces it with $s_1 E(y,k) s_2$. The encryption algorithm used is the Caesar cipher.

This resembles the scenario of instruction-tuning a pre-trained language model, where it already has the ability to replace strings, and rotate tokens, but need to be supervised to capture the meaning of an instruction and follow it correctly.

\begin{figure}[ht]
     \centering
     \begin{subfigure}[h]{.48\textwidth}
     \centering
         \includegraphics[width=\columnwidth]{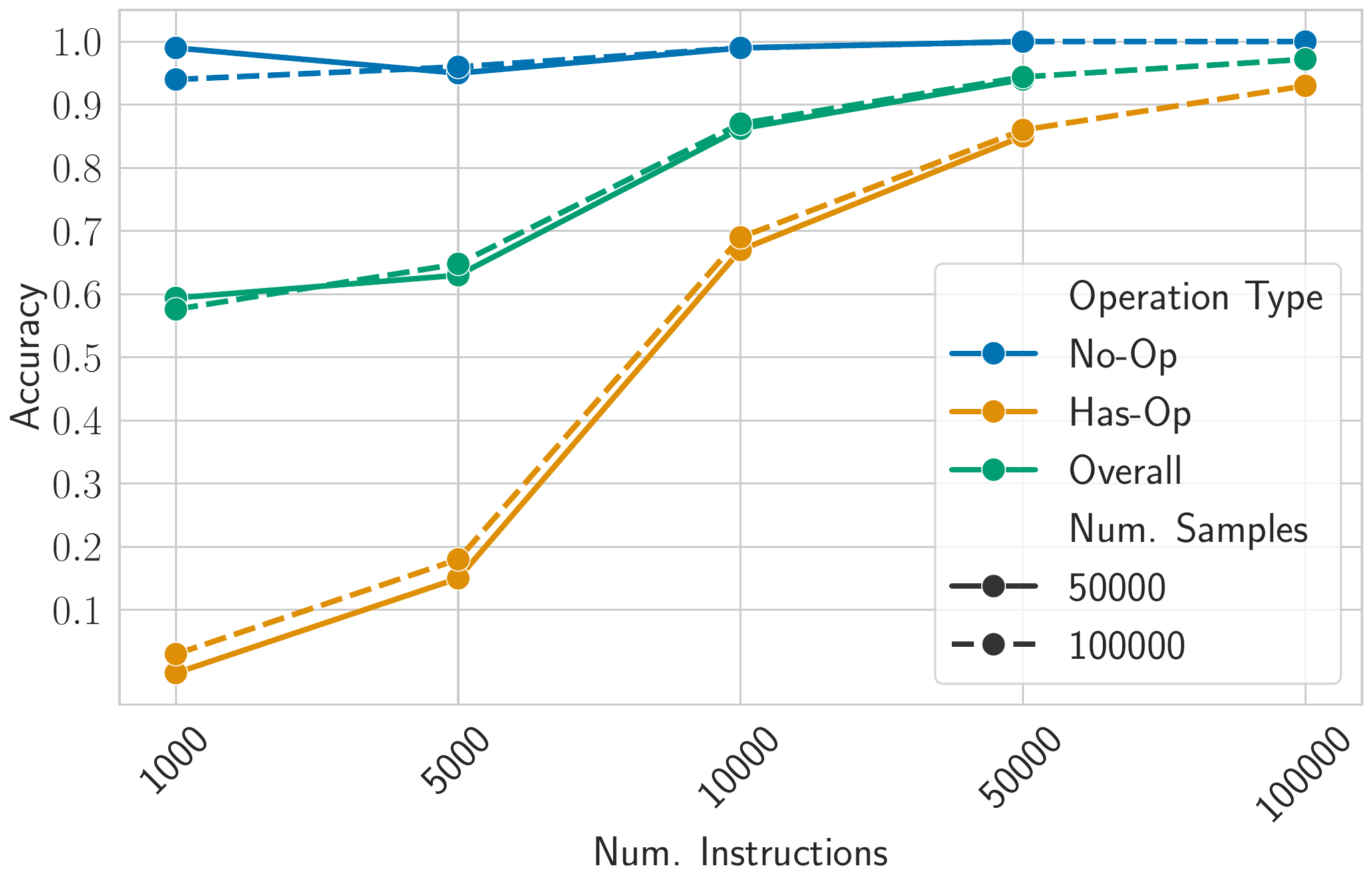}
         \caption{\small Performance of Llama-2 model on the encrypted-rewriting task. We also conducted uniform / non-uniform sub-samplings to half the total sample size at 9000 instructions. Uniform sub-sampling does not harm performance whereas non-uniform subsampling impacts generalization.}
         \label{fig:realworld_perf}
     \end{subfigure}
     \hfill
     \centering
     \begin{subfigure}[h]{.48\textwidth}
     \centering
         \includegraphics[width=\columnwidth]{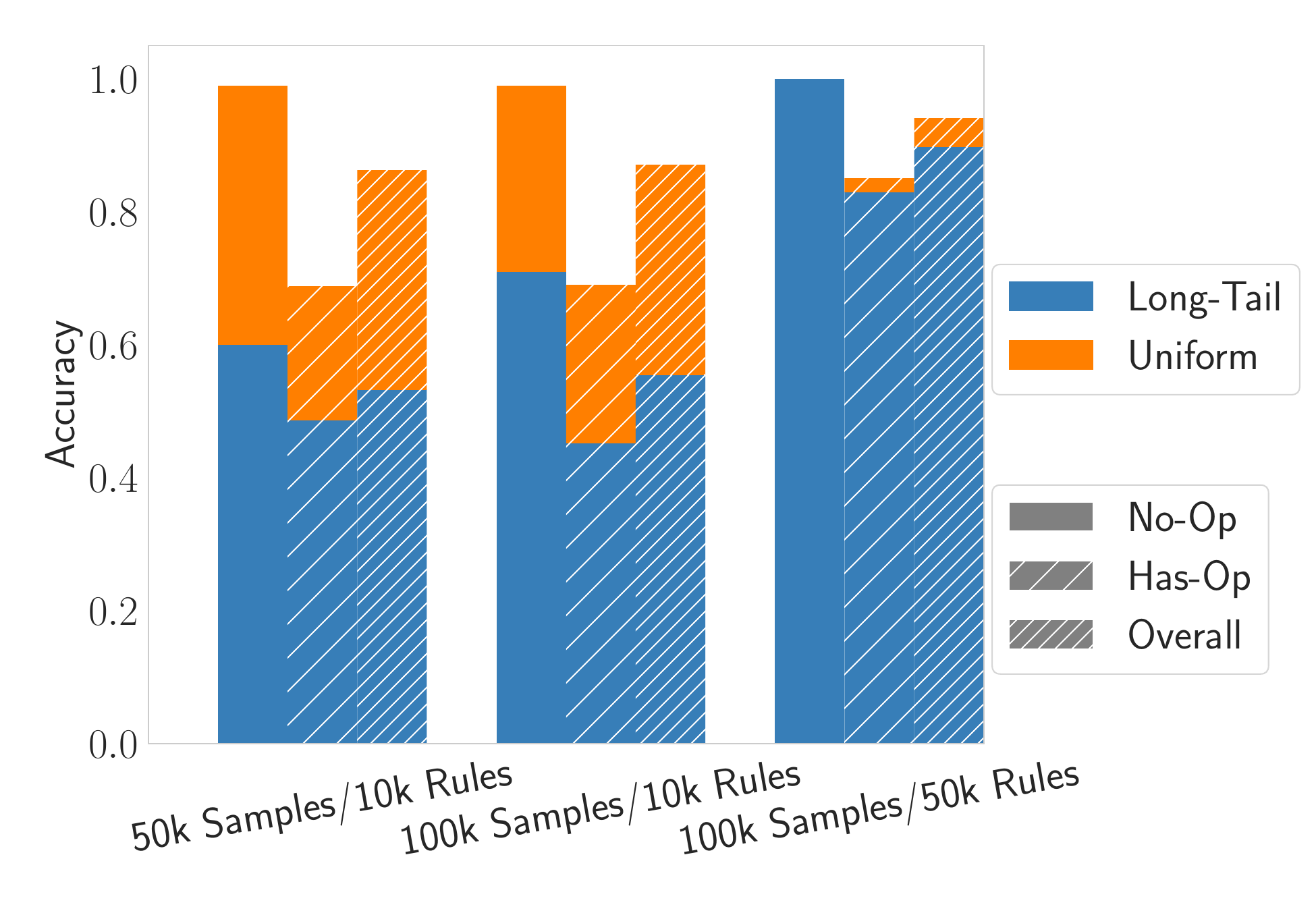}
         \caption{\small Effect of long-tail distribution on encrypted-rewriting experiments. As the number of rules increase, the impact of imbalanced distribution diminishes.}
         \label{fig:long-tail-real-world}
     \end{subfigure}
     \caption{\small Generalization versus the number of instructions during training. }
\end{figure}

We keep two disjoint dictionaries for train and test sets and prompt GPT-3.5-turbo to generate sentences containing words from the dictionary. If the word is in the generated sentence, we randomly sample a replacement and encrypt it with a random key. In no-ops cases, the input should be returned. We generate training sets of $40,000$ sequences and test them on sets of $5,000$ instances - each generated using a distinct word in the test dictionary and again a randomly chosen key. Both sets contain $40\%$ no-ops cases.

We fine-tuned the pre-trained language model (Llama2-7b)~\cite{touvron2023llama} with LoRA~\cite{hu2021lora} with rank 256 and $\alpha$ of 256 till convergence.  Consistent with our earlier observations, the diversity of instructions benefits the model's generalization. With a smaller number of instructions, the pre-trained LLM also only solves no-op cases but cannot correctly perform the ``replace-then-encrypt'' operation (Figure~\ref{fig:realworld_perf}). The impact of long-tailed distributions over rules, in this case, also can be diluted by the diversification of rules appearing in the dataset, as shown in Figure~\ref{fig:long-tail-real-world}.

\section{Effect of Diversification on Real-World Scenario: Code Generation}

Our synthetic experiments suggest that curating datasets with a small number of diversified instructions significantly enhances model performance for instruction following. Diversifying instructions improves the model's ability to handle previously unseen instructions and benefits the model even if the dataset does not cover the entire semantic space, provided the restrictions are not too stringent. Additionally, the model gains from diversification even in long-tailed distributions, where a few instruction classes dominate. This highlights that proper diversification can boost specialized models.

\begin{figure}[ht]
     \centering
     \begin{subfigure}[h]{.32\textwidth}
     \centering
         \includegraphics[width=\columnwidth]{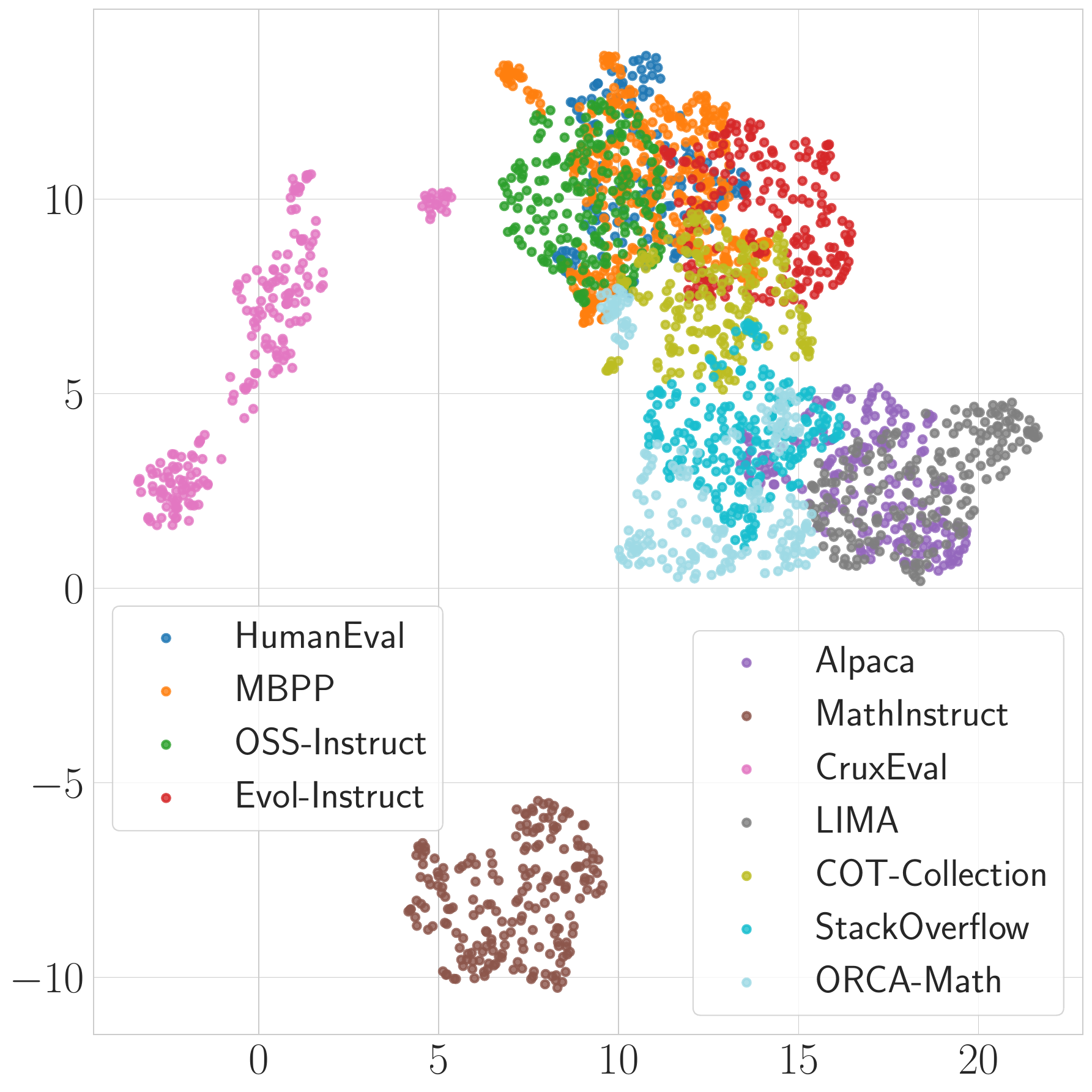}
         \caption{\small Semantic clustering of relevant datasets.}
         \label{fig:emb-all}
     \end{subfigure}
     \hfill
     \centering
     \begin{subfigure}[h]{.32\textwidth}
     \centering
         \includegraphics[width=\columnwidth]{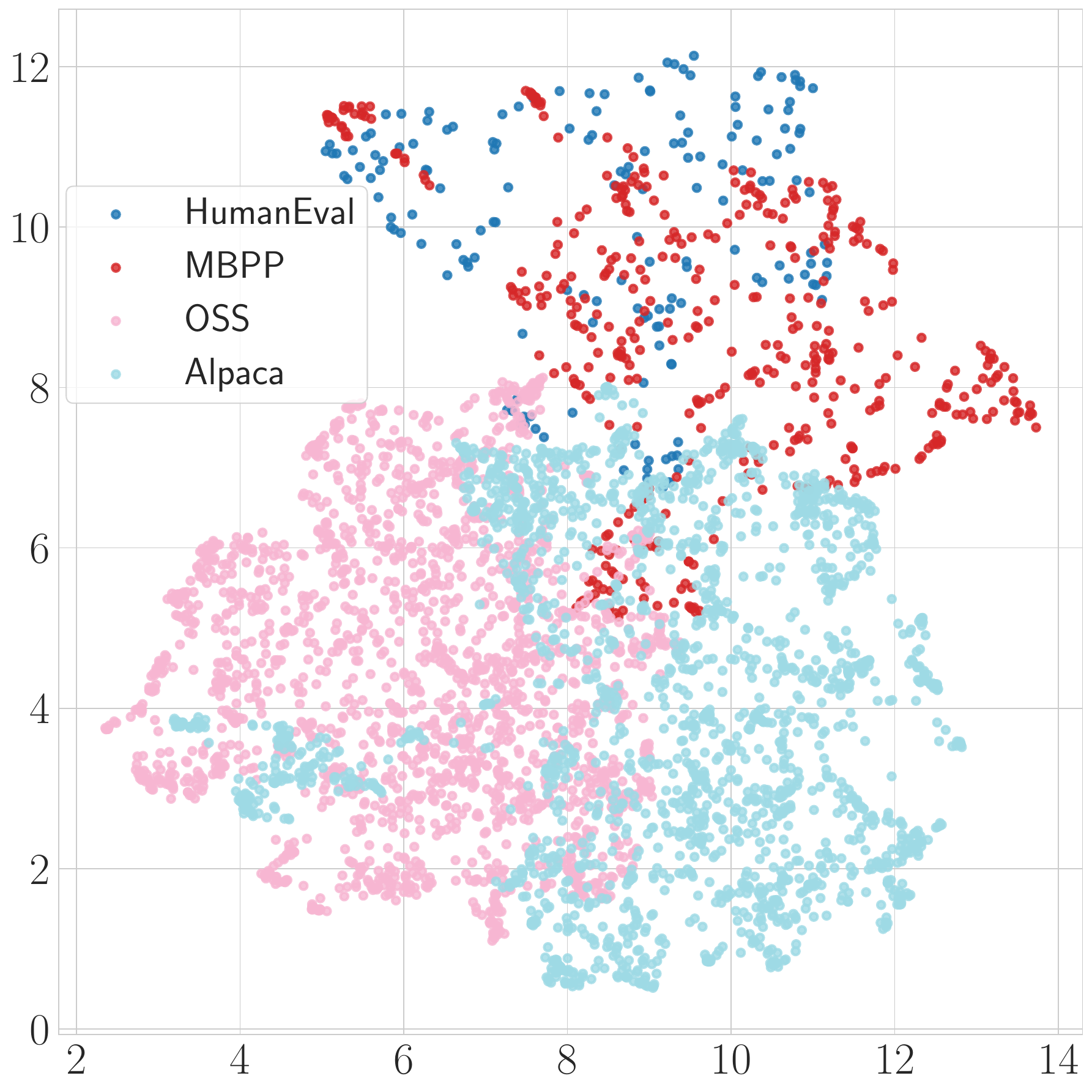}
         \caption{\small OSS-Alpaca mixture and test instructions.}
         \label{fig:emb-train}
     \end{subfigure}
     \hfill
     \centering
     \begin{subfigure}[h]{.32\textwidth}
     \centering
         \includegraphics[width=\columnwidth]{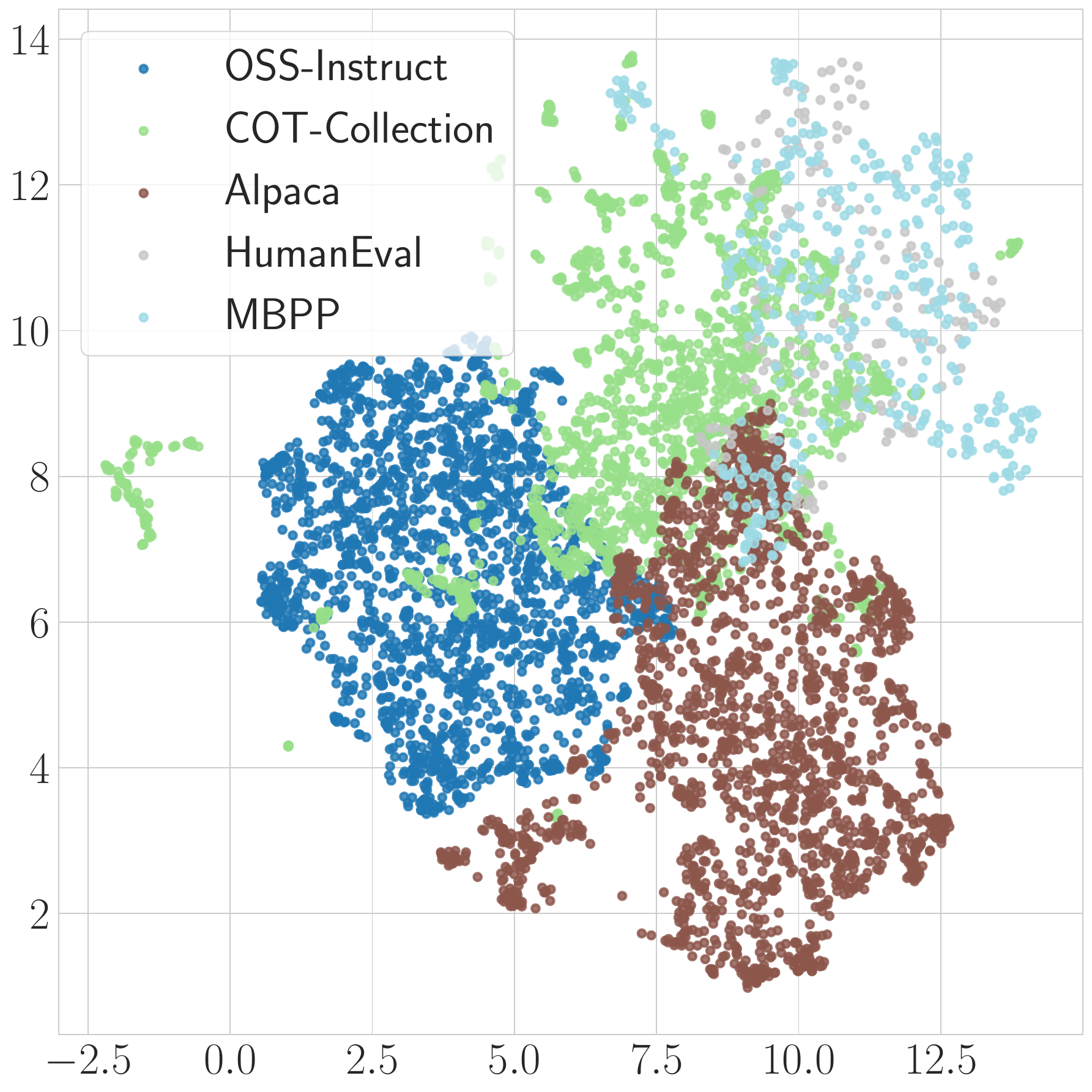}
         \caption{OSS-COT-Alpaca mixture and test instructions.}
         \label{fig:emb-train-cot}
     \end{subfigure}
     \caption{\small Visualization of embedded instructions using OpenAI Embedding model.}
\end{figure}

This suggests that when constructing real-world instruction-tuning datasets for some specific task, it may not be necessarily optimal 
to fine-tune the model entirely on instruction-response pairs of that specific task. In other words, given a fixed data budget, 
\textbf{it might be better if a certain quantity of data from a wider range of task domains is included in the instruction following data, 
instead of exhausting the budget on the data from that single domain}. In addition, prior works ~\cite{dong2024composition} have found that the performance 
on coding and math reasoning tasks consistently grows with data size. We therefore conjecture that an optimal mixture might exist under each setting, depending
on the model, datasets, and number of instances. 

\subsection{Experiments}
We demonstrate how the findings from our earlier investigation can potentially lead to benefits in real-world instruction-following. We experimented with the task of code generation, in which the model re-writes natural language descriptions of a program into code. Besides, this task enjoys a relatively subjective evaluation protocol based on functional correctness. 
By applying our insights on instruction diversification to the domain of program generation, we can effectively assess whether a diversified instruction set enhances the models' performance in handling unseen instructions. 

We evaluate two popular code generation benchmarks, HumanEval~\cite{chen2021humaneval} and MBPP~\cite{austin2021mbpp}, as well as 
EvalPlus -- a widely-adopted augmented version~\cite{liu2023evalplus} of these two datasets. We sample training instances from OSS-Instruct~\cite{wei2023magicoder} dataset for code generation training. OSS-Instruct 
is a synthetic dataset generated using GPT-3.5-turbo and has gone through a sanitizing phase to eliminate data contamination. 
We sample $20,000$ training instances from OSS-Instruct as our baseline instruction-tuning dataset. 
Then we gradually remove code instruction-tuning data and incorporate randomly sampled 
general-domain instruction-tuning data. We used Alpaca~\cite{alpaca}, which is one of the most widely-known instruction datasets for general domain instruction following, diversified across different domains to spread across a wide semantic domain. As shown in figure~\ref{fig:emb-all}, its embedding space overlaps with coding, reasoning, and mathematics domains.

We chose two well-performing pre-trained code LM base models - DeepSeek-Coder-6.7B-Base~\cite{guo2024deepseekcoder} and CodeQwen-7B-Base~\cite{bai2023qwen}. 
\subsection{Instruction Diversity Can Help, But There Is A Price To Pay}
\paragraph{Inclusion of Non-Coding Data} As shown in tables~\ref{tab:deepseek} and ~\ref{tab:codeqwen}, extending the semantics of instruction-tuning data, at a price of even suppressing the number of code generation-specific data points, 
\begin{wrapfigure}{r}{0.5\textwidth}
  \centering
  \includegraphics[width=0.48\textwidth]{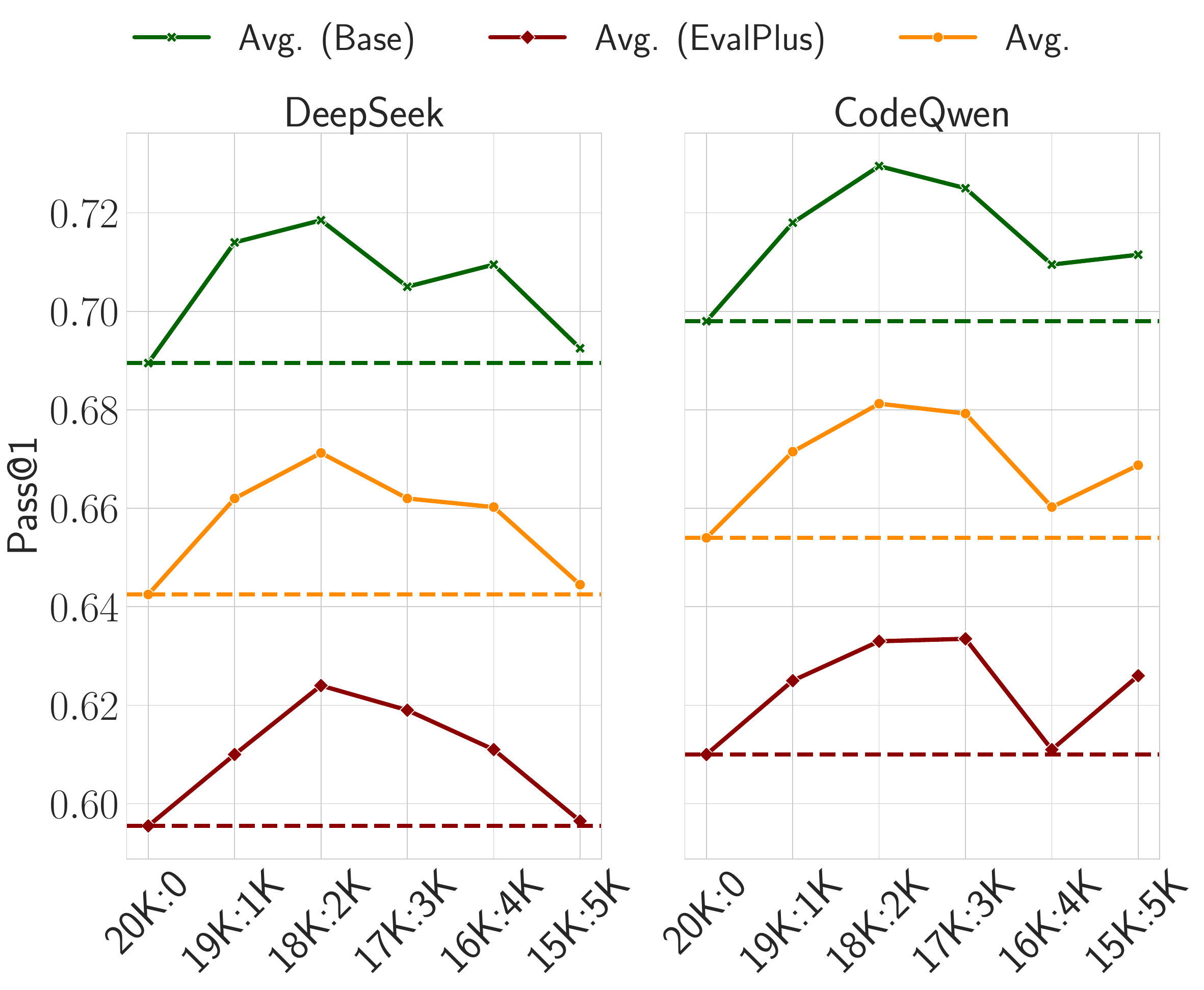} 
         \caption{Trend of Pass@1 with data mixture. The baseline result is marked with dotted lines.}
         \label{fig:trend}
\end{wrapfigure}
consistently brings better overall performance for code-generation tasks, which echos our previous findings from string-rewriting experiments. However, by incorporating general-domain data, we pay a price of less diversification within the code-generation domain. Our results show that the benefit of introducing Alpaca data does not accumulate forever as we trade the diversity within the code generation domain (which is the target domain) with that across different domains. 

This trade-off occurs because while general-domain data can improve the model's ability to understand and generate a wider variety of instructions, it also dilutes the model's specialization in the code-generation domain~\cite{ling2024domainspecilization}. As the proportion of general-domain data increases, the model's focus on the specific patterns and intricacies of code generation diminishes, leading to a plateau or even a decline in performance on code-generation tasks.
Therefore, there exists the best mixture from which the model achieves optimal performance for both models respectively, as shown in Figure~\ref{fig:trend}. 
\begin{table}
\centering
\small
\begin{tabular}{@{}rrrrrr|rrr@{}}
\toprule
\multicolumn{1}{c}{\textbf{\begin{tabular}[c]{@{}c@{}}OSS\\ -Inst.\end{tabular}}} &
  \multicolumn{1}{l}{\textbf{Alpaca}} &
  \multicolumn{1}{l}{\textbf{HumanEval}} &
  \multicolumn{1}{l}{\textbf{HE+}} &
  \multicolumn{1}{l}{\textbf{MBPP}} &
  \multicolumn{1}{l|}{\textbf{MBPP+}} &
  \multicolumn{1}{c}{\textbf{\begin{tabular}[c]{@{}c@{}}Avg. \\ (Base)\end{tabular}}} &
  \multicolumn{1}{c}{\textbf{\begin{tabular}[c]{@{}c@{}}Avg. \\ (+)\end{tabular}}} &
  \multicolumn{1}{l}{\textbf{Avg.}} \\ \midrule
\textbf{20K} &
  \textbf{0} &
  62.2 &
  56.7 &
  75.7 &
  62.4 &
  68.9 &
  59.5 &
  64.2 \\ \midrule
\textbf{18K} &
  \textbf{2K} &
  \cellcolor[HTML]{D9EAD3}68.3 &
  \cellcolor[HTML]{D9EAD3}61.6 &
  \cellcolor[HTML]{FFCCC9}75.4 &
  \cellcolor[HTML]{D9EAD3}63.2 &
  \cellcolor[HTML]{D9EAD3}\textbf{71.9} &
  \cellcolor[HTML]{D9EAD3}\textbf{62.4} &
  \cellcolor[HTML]{D9EAD3}\textbf{67.2} \\
\textbf{17K} &
  \textbf{3K} &
  \cellcolor[HTML]{D9EAD3}64.6 &
  \cellcolor[HTML]{D9EAD3}60.4 &
  \cellcolor[HTML]{D9EAD3}76.4 &
  \cellcolor[HTML]{D9EAD3}63.4 &
  \cellcolor[HTML]{D9EAD3}70.5 &
  \cellcolor[HTML]{D9EAD3}{\textbf{\textit{61.9}}} &
  \cellcolor[HTML]{D9EAD3}\textit{\textbf {66.2}} \\
\textbf{19K} &
  \textbf{1K} &
  \cellcolor[HTML]{D9EAD3}67.1 &
  \cellcolor[HTML]{D9EAD3}59.8 &
  \cellcolor[HTML]{D9EAD3}75.7 &
  \cellcolor[HTML]{FFCCC9}62.2 &
  \cellcolor[HTML]{D9EAD3}{\textbf{\textit{71.4}}} &
  \cellcolor[HTML]{D9EAD3}61.0 &
  \cellcolor[HTML]{D9EAD3}66.2 \\
\textbf{16K} &
  \textbf{4K} &
  \cellcolor[HTML]{D9EAD3}65.2 &
  \cellcolor[HTML]{D9EAD3}58.5 &
  \cellcolor[HTML]{D9EAD3}76.7 &
  \cellcolor[HTML]{D9EAD3}63.7 &
  \cellcolor[HTML]{D9EAD3}71.0 &
  \cellcolor[HTML]{D9EAD3}61.1 &
  \cellcolor[HTML]{D9EAD3}66.0 \\
\textbf{15K} &
  \textbf{5K} &
  \cellcolor[HTML]{D9EAD3}64.6 &
  \cellcolor[HTML]{D9EAD3}57.9 &
  \cellcolor[HTML]{FFCCC9}73.9 &
  \cellcolor[HTML]{FFCCC9}61.4 &
  \cellcolor[HTML]{D9EAD3}69.3 &
  \cellcolor[HTML]{D9EAD3}59.7 &
  \cellcolor[HTML]{D9EAD3}64.5 \\ \bottomrule
\end{tabular}
\caption{\small Pass@1 on DeepSeek-Coder-6.7B. We highlighted the accuracies surpassing baseline result with green and those falling below with red. }
\label{tab:deepseek}
\end{table}
The results echo our earlier conjecture drawn from the string-rewriting experiments. The inclusion of the Alpaca dataset highly
diversifies the distribution of SFT data. 
\begin{table}
\centering
\small
\begin{tabular}{@{}rrrrrr|rrr@{}}
\toprule
\multicolumn{1}{c}{\textbf{\begin{tabular}[c]{@{}c@{}}OSS\\ -Inst.\end{tabular}}} &
  \multicolumn{1}{l}{\textbf{Alpaca}} &
  \multicolumn{1}{l}{\textbf{HumanEval}} &
  \multicolumn{1}{l}{\textbf{HE+}} &
  \multicolumn{1}{l}{\textbf{MBPP}} &
  \multicolumn{1}{l|}{\textbf{MBPP+}} &
  \multicolumn{1}{c}{\textbf{\begin{tabular}[c]{@{}c@{}}Avg. \\ (Base)\end{tabular}}} &
  \multicolumn{1}{c}{\textbf{\begin{tabular}[c]{@{}c@{}}Avg. \\ (+)\end{tabular}}} &
  \multicolumn{1}{l}{\textbf{Avg.}} \\ \midrule
\textbf{20K} &
  \textbf{0} &
  65.9 &
  59.8 &
  73.7 &
  62.2 &
  69.8 &
  61.0 &
  65.4 \\ \midrule
\textbf{18K} &
  \textbf{2K} &
  \cellcolor[HTML]{D9EAD3}69.5 &
  \cellcolor[HTML]{D9EAD3}63.4 &
  \cellcolor[HTML]{D9EAD3}76.4 &
  \cellcolor[HTML]{D9EAD3}63.2 &
  \cellcolor[HTML]{D9EAD3}\textbf{73.0} &
  \cellcolor[HTML]{D9EAD3}\textit{\textbf{63.3}} &
  \cellcolor[HTML]{D9EAD3}\textbf{68.1} \\
  \textbf{17K} &
  \textbf{3K} &
  \cellcolor[HTML]{D9EAD3}68.3 &
  \cellcolor[HTML]{D9EAD3}62.8 &
  \cellcolor[HTML]{D9EAD3}76.7 &
  \cellcolor[HTML]{D9EAD3}63.9 &
  \cellcolor[HTML]{D9EAD3}\textit{\textbf{72.5}} &
  \cellcolor[HTML]{D9EAD3}\textbf{63.4} &
  \cellcolor[HTML]{D9EAD3}\textit{\textbf{67.9}} \\
\textbf{19K} &
  \textbf{1K} &
  \cellcolor[HTML]{D9EAD3}67.7 &
  \cellcolor[HTML]{D9EAD3}61.6 &
  \cellcolor[HTML]{D9EAD3}75.9 &
  \cellcolor[HTML]{D9EAD3}63.4 &
  \cellcolor[HTML]{D9EAD3}{71.8} &
  \cellcolor[HTML]{D9EAD3}62.5 &
  \cellcolor[HTML]{D9EAD3}{67.2} \\
\textbf{15K} &
  \textbf{5K} &
  \cellcolor[HTML]{D9EAD3}67.1 &
  \cellcolor[HTML]{D9EAD3}62.8 &
  \cellcolor[HTML]{D9EAD3}75.2 &
  \cellcolor[HTML]{D9EAD3}62.4 &
  \cellcolor[HTML]{D9EAD3}71.2 &
  \cellcolor[HTML]{D9EAD3}{62.6} &
  \cellcolor[HTML]{D9EAD3}66.9 \\
\textbf{16K} &
  \textbf{4K} &
  \cellcolor[HTML]{FFCCC9}65.2 &
  \cellcolor[HTML]{FFCCC9}58.5 &
  \cellcolor[HTML]{D9EAD3}76.7 &
  \cellcolor[HTML]{D9EAD3}63.7 &
  \cellcolor[HTML]{D9EAD3}71.0 &
  \cellcolor[HTML]{D9EAD3}{61.1} &
  \cellcolor[HTML]{D9EAD3}66.0 \\
\bottomrule
\end{tabular}
\caption{\small Pass@1 with CodeQwen-7B.   We highlighted the accuracies surpassing baseline result with green and those falling below with red. }
\label{tab:codeqwen}
\end{table}

Although the majority of the training data still centers around the code generation domain,
the instructions in the Alpaca dataset span across a larger semantic space. Also, although only a small number of instances 
are included for each since the total number of instructions is large enough and the semantics are diverse enough,
the small number of data does play a part in boosting the model's code generation performance. 
\begin{table}
\small
\begin{tabular}{@{}ccrrrr|rrr@{}}
\toprule

\textbf{Data Mixture} &
  \textbf{Ratio} &
  \textbf{HumanEval} &
  \textbf{HE+} &
  \textbf{MBPP} &
  \textbf{MBPP+} &
  \multicolumn{1}{c}{\textbf{\begin{tabular}[c]{@{}c@{}}Avg. \\ (Base)\end{tabular}}} &
  \multicolumn{1}{c}{\textbf{\begin{tabular}[c]{@{}c@{}}Avg. \\ (+)\end{tabular}}} &
  \multicolumn{1}{c}{\textbf{Avg.}} \\ \midrule
\textbf{O}     & \textbf{-}          & 65.9          & 59.8          & 73.7          & 62.2          & 69.8          & 61.0          & 65.4          \\ \midrule
\textbf{O+A}   & \textbf{18:2}       & 69.5          & \textbf{63.4} & 76.4          & 63.2          & 73.0          & 63.3          & 68.1          \\
\textbf{O+C+A} & \textbf{18:1:1}     & \textbf{68.9} & \textbf{63.4} & \textbf{77.4} & \textbf{64.2} & \textbf{73.2} & \textbf{63.8} & \textbf{68.5} \\ \midrule
\textbf{O+A}   & \textbf{16:4}       & 65.2          & 58.5          & 76.7          & 63.7          & 71.0          & 61.1          & 66.0          \\
\textbf{O+C+A} & \textbf{16:2:2}     & \textbf{68.3} & \textbf{61.6} & \textbf{77.2} & \textbf{63.7} & \textbf{72.8} & \textbf{62.7} & \textbf{67.7} \\ \midrule
\textbf{O+A}   & \textbf{17:3}       & \textbf{68.3} & \textbf{62.8} & \textbf{76.7} & \textbf{63.9} & \textbf{72.5} & \textbf{63.4} & \textbf{67.9} \\
\textbf{O+C+A} & \textbf{17:1.5:1.5} & \textbf{68.3} & 61.6          & 75.4          & 62.9          & 71.9          & 62.3          & 67.1          \\ \bottomrule
\end{tabular}
\caption{\small CodeQwen-7B-Base trained with mixture of OSS-Instruct, CoT Collection and Alpaca.\textbf{O} stands for OSS-Instruct, \textbf{C} stands for CoT-Collection and \textbf{A} stands for Alpaca. The best results for each ratio are boldfaced.}
\label{tab:codeqwen-3way}
\end{table}
\paragraph{Diversification Across Domains}

As found in Section~\ref{sec:diversity_semantics}, diversification across instruction semantics is yet another important dimension. As illustrated in Figure ~\ref{fig:emb-all} each instruction-tuning dataset cluster around a specific subspace of semantics, a further variety among instruction semantics could be obtained by further incorporating other datasets that cover a different space. 
Since the Alpaca dataset was mainly curated to build language model assistants that are optimized for interaction with humans, we consider expanding the semantics domain further by additionally incorporating data that requires more complex, multi-hop reasoning from the CoT-Collection dataset~\cite{kim2023cot} which contains reasoning problems and chain-of-thought style responses. 

As shown in Figure~\ref{tab:codeqwen-3way}, we observe the CoT data indeed is complementary in its semantics space to the OSS-Alpaca mixture. 
To investigate the effect of diversification across semantics domains, we trained the models with various OSS-Instruct v.s. non-coding ratios and benchmarked their performances against previous results. The primary goal of this experiment is to explore the benefits of further extending the semantics domain rather than finding the optimal mixture. Therefore, we used equal numbers of instances from the Alpaca and CoT-Collection datasets to rule out additional confounding factors.

The results obtained using CodeQwen are presented in Table~\ref{tab:codeqwen-3way}. Consistent with our findings from string replacement, the model trained on the optimal ratio within the 3-way mixture outperforms the one trained on the optimal OSS-Instruct + Alpaca mixture. This outcome reinforces the advantages of cross-domain diversification in enhancing model performance. However, it is important to note that this performance boost may not be consistent across all splits. When comparing the mix ratios of $OSS:CoT
=17:1.5:1.5$ and $OSS
=17:3$, we observe a performance decline when CoT data is included.  We conjecture that the effectiveness of diversification is influenced by the specific mix ratio between the datasets used.

\begin{table}
\small
\begin{tabular}{@{}lllcccccccc@{}}
\toprule
\textbf{Mix}   & \textbf{Ratio}          & \textbf{\begin{tabular}[c]{@{}l@{}}Total\\ \#Data\end{tabular}} & \textbf{HE}                  & \textbf{HE+}                 & \textbf{MBPP}                & \textbf{MBPP+}               & \textbf{\begin{tabular}[c]{@{}c@{}}Avg\\ (Base)\end{tabular}} & \textbf{\begin{tabular}[c]{@{}c@{}}Avg\\ (+)\end{tabular}} & \textbf{Avg}                          & \textbf{Rel.↑}                       \\ \midrule
\textbf{O}     & \textbf{-}              & \textbf{75K}                                                    & 66.5                         & 61.6                         & 75.4                         & 61.9                         & 71.0                                                          & 61.8                                                       & 66.4                                  & -                                    \\
\textbf{O+A}   & \textbf{60:15}          & \textbf{75K}                                                    & \cellcolor[HTML]{D9EAD3}67.1 & \cellcolor[HTML]{F4CCCC}59.8 & \cellcolor[HTML]{D9EAD3}77.0 & \cellcolor[HTML]{D9EAD3}64.8 & \cellcolor[HTML]{D9EAD3}72.1                                  & \cellcolor[HTML]{D9EAD3}62.3                               & \cellcolor[HTML]{D9EAD3}67.2          & \cellcolor[HTML]{D9EAD3}1.2          \\
\textbf{O+A}   & \textbf{67.5:7.5}       & \textbf{75K}                                                    & \cellcolor[HTML]{D9EAD3}68.9 & \cellcolor[HTML]{D9EAD3}61.6 & \cellcolor[HTML]{D9EAD3}76.2 & \cellcolor[HTML]{D9EAD3}63.8 & \cellcolor[HTML]{D9EAD3}72.6                                  & \cellcolor[HTML]{D9EAD3}62.7                               & \cellcolor[HTML]{D9EAD3}67.6          & \cellcolor[HTML]{D9EAD3}1.9          \\
\textbf{O+A}   & \textbf{72.5:2.5}       & \textbf{75K}                                                    & \cellcolor[HTML]{D9EAD3}68.3 & \cellcolor[HTML]{F4CCCC}61.0 & \cellcolor[HTML]{D9EAD3}76.2 & \cellcolor[HTML]{D9EAD3}64.3 & \cellcolor[HTML]{D9EAD3}72.3                                  & \cellcolor[HTML]{D9EAD3}62.7                               & \cellcolor[HTML]{D9EAD3}67.5          & \cellcolor[HTML]{D9EAD3}1.6          \\
\textbf{O+A}   & \textbf{74:1}           & \textbf{75K}                                                    & \cellcolor[HTML]{D9EAD3}69.5 & \cellcolor[HTML]{D9EAD3}62.2 & \cellcolor[HTML]{D9EAD3}76.2 & \cellcolor[HTML]{D9EAD3}64.3 & \cellcolor[HTML]{D9EAD3}72.9                                  & \cellcolor[HTML]{D9EAD3}\textit{63.3}                      & \cellcolor[HTML]{D9EAD3}\textit{68.1} & \cellcolor[HTML]{D9EAD3}\textit{2.5} \\ \midrule
\textbf{O+C+A} & \textbf{60:7.5:7.5}     & \textbf{75K}                                                    & \cellcolor[HTML]{F4CCCC}64.6 & \cellcolor[HTML]{F4CCCC}59.1 & \cellcolor[HTML]{D9EAD3}76.2 & \cellcolor[HTML]{D9EAD3}63.8 & \cellcolor[HTML]{F4CCCC}70.4                                  & \cellcolor[HTML]{F4CCCC}61.5                               & \cellcolor[HTML]{D9EAD3}65.9          & \cellcolor[HTML]{F4CCCC}-0.6         \\
\textbf{O+C+A} & \textbf{67.5:3.25:3.25} & \textbf{75K}                                                    & \cellcolor[HTML]{D9EAD3}66.5 & \cellcolor[HTML]{D9EAD3}62.2 & \cellcolor[HTML]{D9EAD3}76.2 & \cellcolor[HTML]{D9EAD3}64.3 & \cellcolor[HTML]{D9EAD3}71.4                                  & \cellcolor[HTML]{D9EAD3}63.3                               & \cellcolor[HTML]{D9EAD3}67.3          & \cellcolor[HTML]{D9EAD3}1.4          \\
\textbf{O+C+A} & \textbf{72.5:1.25:1.25} & \textbf{75K}                                                    & \cellcolor[HTML]{D9EAD3}68.3 & \cellcolor[HTML]{D9EAD3}62.2 & \cellcolor[HTML]{D9EAD3}77.8 & \cellcolor[HTML]{D9EAD3}65.1 & \cellcolor[HTML]{D9EAD3}73.1                                  & \cellcolor[HTML]{D9EAD3}\textbf{63.7}                      & \cellcolor[HTML]{D9EAD3}\textbf{68.4} & \cellcolor[HTML]{D9EAD3}\textbf{3.0} \\
\textbf{O+C+A} & \textbf{74:0.5:0.5}     & \textbf{75K}                                                    & \cellcolor[HTML]{D9EAD3}66.5 & \cellcolor[HTML]{F4CCCC}61.0 & \cellcolor[HTML]{D9EAD3}76.5 & \cellcolor[HTML]{D9EAD3}63.2 & \cellcolor[HTML]{D9EAD3}71.5                                  & \cellcolor[HTML]{D9EAD3}62.1                               & \cellcolor[HTML]{D9EAD3}66.8          & \cellcolor[HTML]{D9EAD3}0.7          \\ \bottomrule
\end{tabular}
\caption{Comparison With MagiCoder-DS-6.7B~\cite{wei2023magicoder}. We demonstrated that one could achieve higher performances by means of diversification. }
\label{tab:magicoder}
\end{table}

\subsection{Beating Open Source Code-LMs With Its Own Data (+ Diversification)}

As yet a further step to demonstrate the benefit of diversification, we train a code language model by applying the same ideas of diversification to the whole OSS-Instruct dataset and compare the performance against the model trained on it: MagiCoder-DS-6.7B~\cite{wei2023magicoder}. We train all the trainable weights (instead of LoRA) in this set of experiments. As shown in Table~\ref{tab:magicoder}, appropriate diversification still leads to performance advantages. The observations are also consistent with our earlier experiments for code generation and string replacement, where a small fraction of non-coding data can drive performance gain, and that diversifying across multiple domains can push the performance even higher.

\section{Conclusion and Limitations}
Through our symbolic experiments, we have shown that language models only generalize to instructions unseen during training when trained on a large and diverse set of instructions. For a fixed data budget, instruction diversity outweighs better illustration (i.e. more examples) for each instruction. These observations apply not only to the number of different instructions in the dataset but also to their semantic diversity. The negative effect of an unbalanced distribution of examples can be counteracted by a larger number of instructions in the training set. Putting all these together, this implies the possibility of improving the model's performance on some task by incorporating a small portion of data from other task domains. We demonstrated that diversity can bring benefits for real-world instruction following using the example scenario of code generation, and provided insights into effective diversification for better task performances.

\paragraph{Limitations} We did not propose an algorithm to find the best data mixture including domains and ratios for task-specific instruction-tuning but differed the methods to search for the optimal strategy of diversification for future work.
\section*{Acknowledgement}
We thank Chi Han, Jialiang Xu, Yifeng Ding and Yuchen Li for providing suggestions and valuable feedback on this project. 
\newpage
\bibliography{references}
\bibliographystyle{plain}
\newpage
\appendix
\section{Complement on Markov algorithms}
\label{app:markov}

Markov algorithms~\cite{markov54} are ordered sets of rewrite rules, operating on sequences of symbols in a fixed alphabet $\mathcal U$. A sequence $S$ is processed by applying the first rewrite applicable to $S$, at the leftmost position if several exist: i.e. the rewrite rule $ss \to tr$ transforms the sequence $S=mississipi$ into $S'=mitrissipi$. The algorithm is then applied to $S'$, and the process is repeated until either no rules apply, and the algorithm is said to be \emph{blocked}, or a special rule, called a \emph{stop rule} is invoked, and the algorithm terminates and returns the final rewritten sequence.

Specifically, the algorithm uses and alphabet $\mathcal A$, which includes the alphabet $\mathcal U$ used buy the sequences to be processed (henceforth, small case latin letters), a set of additional symbols (henceforth, the small case greek letters $\{\alpha, \beta \dots \}$, and a special symbol $\cdot$ indicating a stop rule.

For instance, we could define the following algorithm, with $\mathcal U=\{a,b\}$, and $\mathcal A=\{a,b,\alpha,\beta,\cdot\}$, and the rules

\begin{eqnarray} 
\alpha x &\to& x \alpha \beta x \\
\beta xy &\to& y \beta x \\
\alpha \beta x &\to& x \alpha \\
\alpha &\to& \cdot \\
 &\to& \alpha
\end{eqnarray}
where $x$ and $y$ stand for any letter $a$ or $b$. This will transform any sequence of $a$ and $b$ into a concatenation of the sequence and its reverse. Applied on $abb$, the algorithm will perform the following rewrites: 

\begin{align*} 
abb &\to \alpha abb &&(\text{by }5)\\
\alpha abb  &\to a \alpha \beta abb &&(\text{by }1)\\ 
a \alpha \beta abb &\to a \alpha b \beta ab &&(\text{by }2)\\
a \alpha b \beta ab &\to a b \alpha \beta b \beta ab &&(\text{by }1)\\
 a b \alpha b \beta b \beta ab &\to  a b \alpha \beta bb \beta a &&(\text{by }2)\\
 a b \alpha \beta bb \beta a&\to a b \alpha b \beta b \beta a &&(\text{by }2)\\
a b \alpha b \beta b \beta a &\to abb \alpha \beta b \beta b \beta a &&(\text{by }1)\\ 
abb \alpha \beta b \beta b \beta a &\to abb b \alpha \beta b \beta a &&(\text{by }3)\\
abb b \alpha \beta b \beta a &\to abbbb \alpha \beta a &&(\text{by }3)\\ 
abbbb \alpha \beta a &\to abbbba \alpha &&(\text{by }3)\\
abbbba \alpha &\to abbbba &&(\text{by }4)
\end{align*}

Since rule $4$ is a stop rule, the algorithm terminates and returns $abbbba$.

Judicious introduction of additional (greek) letters allows one to compose Markov algorithms, effectively writing complex programs. Any effective process (i.e. finite computation) can be represented as a Markov algorithm (this is Markov's thesis).

\section{Experimental set-up}
\label{task}
\subsection{Model and Training}
In rewrite experiments, we train GPT-2 models~\cite{Radford2019gpt2}, a decoder-only transformer-based architecture, with $6$ layers, $256$ dimensions and $4$ attention heads from scratch, on a generated instruction-tuning dataset using standard supervised fine-tuning approach.  We use the AdamW optimizer, a learning rate of $10^{-3}$, and linear scheduling. All models are trained for 50 epochs. For the encrypted-rewriting task, we LoRA fine-tuned Llama-2 models with a learning rate of 1e-5, batch size 64, gradient accumulation step 1, and 8-bit quantization. The model takes about 2000 steps to converge. For coding experiments, we trained the model with a learning rate of 1e-5, batch size 4, and gradient accumulation step 1, 8-bit quantization for 3 epochs with a maximum length of 768. The models are trained and inferenced on 1 Nvidia A40 GPU. We used greedy decoding for all experiments. 

\subsection{Data Generation}
\paragraph{Synthetic Experiment}
Except for the diversity of semantics experiment, the results we reported in the main paper are obtained from an input length of 50 and a pattern length of 20.
To validate the generality of our findings, we conducted experiments on various input sizes \{50, 100, 200\} and, correspondingly, pattern lengths \{20,40,50\}.

In the diversity of semantics experiment, we used an input length of 500 and a pattern length of 60. We strictly restricted the sub-strings to look for and to replace them with both to be unseen during testing. 
\paragraph{Code Generation}
We downloaded the data from the official Huggingface Datasets Repos.

\end{document}